\documentclass[journal,transmag]{IEEEtran}

\usepackage[english]{babel}
\usepackage[utf8x]{inputenc}
\usepackage[T1]{fontenc}
\usepackage{enumitem}
\usepackage[a4paper,top=3cm,bottom=2cm,left=3cm,right=3cm,marginparwidth=1.75cm]{geometry}

\usepackage{amsmath}
\usepackage{graphicx}
\usepackage{microtype}
\usepackage[colorlinks=true, allcolors=blue]{hyperref}

\usepackage{subcaption}
\usepackage{array}

\title{Procedural Content Generation via Machine Learning (PCGML)}
\author{\IEEEauthorblockN{Adam Summerville\IEEEauthorrefmark{1},
Sam Snodgrass\IEEEauthorrefmark{2},
Matthew Guzdial\IEEEauthorrefmark{3}, 
Christoffer Holmgård\IEEEauthorrefmark{4},\\
Amy K. Hoover\IEEEauthorrefmark{5},
Aaron Isaksen\IEEEauthorrefmark{6},
Andy Nealen\IEEEauthorrefmark{6}, and
Julian Togelius\IEEEauthorrefmark{6},
}
\IEEEauthorblockA{\IEEEauthorrefmark{1}Department of Computational Media, University of California, Santa Cruz, CA 95064, USA}
\IEEEauthorblockA{\IEEEauthorrefmark{2}College of Computing and Informatics, Drexel University, Philadelpia, PA 19104, USA}
\IEEEauthorblockA{\IEEEauthorrefmark{3}School of Electrical and Computer Engineering,
Georgia Institute of Technology, Atlanta, GA 30332, USA}
\IEEEauthorblockA{\IEEEauthorrefmark{4}Duck and Cover Games ApS, 1311 Copenhagen K, Denmark}
\IEEEauthorblockA{\IEEEauthorrefmark{5}College of Arts, Media and Design, Northeastern University, Boston, MA 02115, USA}
\IEEEauthorblockA{\IEEEauthorrefmark{6}Department of Computer Science and Engineering, New York University, Brooklyn, NY 11201, USA}
\IEEEauthorblockA{emails: asummerv@ucsc.edu, sps74@drexel.edu, mguzdial3@gatech.edu, christoffer@holmgard.org,\\
amy.hoover@gmail.com, aisaksen@appabove.com, nealen@nyu.edu, julian@togelius.com}
}% <-this % stops an unwanted space
\IEEEtitleabstractindextext{%
\begin{abstract}
This survey explores \emph{Procedural Content Generation via Machine Learning} (PCGML), defined as the generation of game content using machine learning models trained on existing content. As the importance of PCG for game development increases, researchers explore new avenues for generating high-quality content with or without human involvement; this paper addresses the relatively new paradigm of using machine learning (in contrast with search-based, solver-based, and constructive methods). We focus on what is most often considered functional game content such as platformer levels, game maps, interactive fiction stories, and cards in collectible card games, as opposed to cosmetic content such as sprites and sound effects. In addition to using PCG for autonomous generation, co-creativity, mixed-initiative design, and compression, PCGML is suited for repair, critique, and content analysis because of its focus on modeling existing content. We discuss various data sources and representations that affect the generated content.  Multiple PCGML methods are covered, including neural networks: long short-term memory (LSTM) networks, autoencoders, and deep convolutional networks; Markov models: $n$-grams and multi-dimensional Markov chains; clustering; and matrix factorization.  Finally, we discuss open problems in PCGML, including learning from small datasets, lack of training data, multi-layered learning, style-transfer, parameter tuning, and PCG as a game mechanic.

\end{abstract}

}
\begin{document}
\maketitle
\IEEEdisplaynontitleabstractindextext

\section{Introduction}

Procedural content generation (PCG), the creation of game content through algorithmic means, has become increasingly prominent within both game development and technical games research. 
It is employed to increase replay value, reduce production cost and effort, to save storage space, or simply as an aesthetic in itself. 
Academic PCG research addresses these challenges, but also explores how PCG can enable new types of game experiences, including games that can adapt to the player. Researchers also address challenges in computational creativity and ways of increasing our understanding of game design through building formal models~\cite{shaker2016procedural}.

In the games industry, many applications of PCG are what could be called ``constructive'' methods, using grammars or noise-based algorithms to create content in a pipeline without evaluation. 
Many other techniques use either search-based methods~\cite{togelius2011search} (for example using evolutionary algorithms) or solver-based methods~\cite{smith2011answer} to generate content in settings that maximize objectives and/or preserve constraints. 
What these methods have in common is that the algorithms, parameters, constraints, and objectives that create the content are in general hand-crafted by designers or researchers. While it is common to examine existing game content for inspiration, machine learning methods have far less commonly been used to extract data from existing game content in order to create more content.

Concurrently, there has been an explosion in the use of machine learning to train models based on datasets~\cite{montgomery2016machine}. 
In particular, the resurgence of neural networks under the name \emph{deep learning} has precipitated a massive increase in the capabilities and application of methods for learning models from big data~\cite{schmidhuber2015deep,goodfellow2016deep}. Deep learning has been used for a variety of tasks in machine learning, including the generation of content. For example, \emph{generative adversarial networks} have been applied to generating artifacts such as images, music, and speech~\cite{goodfellow2014generative}. 
But many other machine learning methods can also be utilized in a generative role, including $n$-grams, Markov models, autoencoders, and others \cite{boulanger2012modeling,fine1998hierarchical,gregor2015draw}. The basic idea is to train a model on instances sampled from some distribution, and then use this model to produce new samples. 

This paper is about the nascent idea and practice of generating game content from machine-learned models.  We define \emph{Procedural Content Generation via Machine Learning} (abbreviated PCGML) as the generation of game content by models that have been trained on existing game content. The difference to search-based~\cite{togelius2011search} and solver-based~\cite{smith2011answer,smith2011tanagra} PCG is that while the latter approaches might use machine-learned models (e.g. trained neural networks) for content \emph{evaluation}, the content generation happens through search in \emph{content space}; in PCGML, the content is generated \emph{directly} from the model. By this we mean that the output of a machine-learned model (given inputs that are either drawn from a random distribution or that represent partial or previous game content) is itself interpreted as content, which is not case in search-based PCG~\footnote{As with any definition, there are corner cases. For exmaple, the Functional Scaffolding approach to generating levels discussed later in this paper can be described as both search-based PCG and PCGML.}. We can further differentiate PCGML from experience-driven PCG~\cite{yannakakis2011experience} through noting that the learned models are models of game content, not models of player experience, behavior or preference. Similarly, learning-based PCG~\cite{roberts2015learning} uses machine learning in several roles, but not for modeling content per se.

The content models could be of many different kinds and trained using very different training algorithms, including neural networks, probabilistic models, decision trees, and others.  The generation could be partial or complete, autonomous, interactive, or guided.  The content could be almost anything in a game, such as levels, maps, items, weapons, quests, characters, rules, etc.

This paper focuses on game content that is directly related to game mechanics. In other words, we focus on \emph{functional} rather than \emph{cosmetic} game content. We define functional content as artifacts that, if they were changed, could alter the in-game effects of a sequence of player actions. The main types of cosmetic game content that we exclude are textures and sound, as those do not directly impact the effects of in-game actions the way levels or rules do in most games, and there is already much research on the generation of such content outside of games~\cite{wei2009state,ren2013example}. This is not a value judgment, and cosmetic content is extremely important in games; however, it is not the focus of this paper. 
%Matthew added
Togelius et al. \cite{togelius2011search} previously defined a related categorization with the terms necessary and optional. We note that while there exists some overlap between necessary and functional, it is possible to have optional functional content (e.g., optional levels) and necessary cosmetic content (e.g., the images and sound effects of a player character).

It is important to note a key difference between game content generation and procedural generation in many other domains: most game content has strict structural constraints to ensure playability. These constraints differ from the structural constraints of text or music because of the need to play games in order to experience them.
Where images, sounds, and in many ways also text can be consumed statically, games are dynamic and must be evaluated through interaction that requires non-trivial effort---in Aarseth's terminology, games are \emph{ergodic media}~\cite{aarseth1997cybertext}.
A level that structurally prevents players from finishing it is not a good level, even if it's visually attractive; a strategy game map with a strategy-breaking shortcut will not be played even if it has interesting features; a game-breaking card in a collectible card game is merely a curiosity; and so on. Thus, the domain of game content generation poses different challenges from that of other generative domains. Of course, there are many other types of content in other domains which pose different, and in some sense more difficult challenges, such as lifelike and beautiful images or evocative musical pieces; however, in this paper we focus on the challenges posed by game content by virtue of its necessity for interaction.

The remainder of this paper is structured as follows. Section~\ref{sec:use-cases} describes the various use cases for PCGML, including various types of generation and uses of the learned models for purposes that are not strictly generative. Section~\ref{sec:data-sources} discusses the key problem of data acquisition and the recurring problem of small datasets.  Section~\ref{sec:Methods} includes a large number of examples of PCGML approaches. As we will see, there is already a large diversity of methodological approaches, but only a limited number of domains have been attempted.  In Section~\ref{sec:open-problems}, we outline a number of important open problems in the research and application of PCGML.

\section{Use Cases for PCGML}
\label{sec:use-cases}
Procedural Content Generation via Machine Learning shares many uses with other forms of PCG: in particular, autonomous generation, co-creation/mixed initiative design, and data compression.  However, because it has been trained on existing content, it can extend into new use areas, such as repair and critique/analysis of new content.

\subsection{Autonomous Generation} The most straightforward application of PCGML is \emph{autonomous PCG}: the generation of complete game artifacts without human input at the time of generation. Autonomous generation is particularly useful when online content generation is needed, such as in rogue-like games.% which require runtime level generation.

PCGML is well-suited for autonomous generation because the input to the system can be examples of representative content specified in the content domain.  With search-based PCG using a generate-and-test framework, a programmer must specify an algorithm for generating the content and an evaluation function that can validate the fitness of the new artifact~\cite{togelius2010search}.  However, designers must use a different domain (code) from the output they wish to generate.  With PCGML, a designer can create a set of representative artifacts in the target domain as a model for the generator, and then the algorithm can generate new content in this style. 
PCGML avoids the complicated step of experts having to codify their design knowledge and intentions.

\subsection{Co-creative and Mixed-initiative Design}
A more compelling use case for PCGML is AI-assisted design, where a human designer and an algorithm work together to create content. This approach has previously been explored with other methods such as constraint satisfaction algorithms and evolutionary algorithms~\cite{shaker2013ropossum,liapis2013sentient,smith2011tanagra}. 

Again, because the designer can train the machine-learning algorithm by providing examples in the target domain, the designer is ``speaking the same language'' the algorithm requires for input and output.  This has the potential to reduce frustration, user error, user training time, and lower the barrier to entry because a programming language is not required to specify generation or acceptance criteria.

PCGML algorithms are provided with example data, and thus are suited to auto-complete game content that is partially specified by the designer.  Within the image domain, we have seen work on image \emph{inpainting}, where a neural network is trained to complete images where parts are missing~\cite{guillemot2014image}. Similarly, machine learning methods could be trained to complete partial game content.% with parts missing.

\subsection{Repair}
With a library of existing representative content, PCGML algorithms can identify areas that are not playable (e.g., if an unplayable level or impossible rule set has been specified) and offer suggestions for how to fix them. Summerville and Mateas \cite{summerville2016mariostring} use a special tile that represents where an AI would choose to move the player in their training set, to bias the algorithm towards generating playable content; the system inherently has learned the difference between passable and impassable terrain.  Jain et. al. \cite{jain2016autoencoders} used a sliding window and an autoencoder to repair illegal level segments -- because they did not appear in the training set, the autoencoder replaced them with a nearby window seen during training.

\subsection{Recognition, Critique, and Analysis}
A use case for PCGML that sets it apart from other PCG approaches is its capacity for recognition, analysis, and critique of game content. Given the basic idea of PCGML is to train some kind of model on sets of existing game content, these models could be applied to analyzing other game content, whether created by an algorithm, players, or designers. 

Previous work has used supervised training to predict properties of content \cite{yannakakis2013player,guzdial2016deep,summerville2017understanding}, but PCGML enables new approaches operating in an unsupervised manner.  Encoding approaches compress the content to an encoded state that can then be analyzed in further processes, such as determining which type of level a piece of content comes from \cite{jain2016autoencoders} or which levels from one game are closest to the levels from a different game \cite{snodgrass2016approach}. %(e.g., which levels from \textit{Kid Icarus} are closest to the levels from \textit{Super Mario Bros.}).  

These learned representations are a byproduct of the generation process, and future work could be used to automatically evaluate game content, as is already done within many applications of search-based PCG, and potentially be used with other generative methods. They also have the potential to identify uniqueness, for example by noting how frequently a particular pattern appears in the training set, or judging how related a complete content artifact is to an existing set.

%For example, by supervised training on sets of game content artifacts and player response data, machine learning models can be trained to classify game content into various types, such as broken or whole, underworld or overworld, etc. They can be trained to predict quantities such as difficulty, completion time, or emotional tenor . They also have the potential to identify uniqueness, for example by noting how frequently a particular pattern appears in the training set, or judging how related a complete content artifact is to an existing set. Being able to identify which user-created levels are truly novel, and then enhance them further so as to fit with the aesthetics of the game and repair broken parts, could be very useful. Such models could be used to automatically evaluate game content, as is already done within many applications of search-based PCG, and potentially be used with other generative methods. 

\subsection{Data Compression} One of the original motivations for PCG, particularly in early games such as \emph{Elite}~\cite{elite}, was data compression. There was not enough space on disk for the game universe. The same is true for some of today's games such as \textit{No Man's Sky}~\cite{nomanssky}. The compression of game data into fewer dimensions through machine learning could allow more efficient game content storage. By exploiting the regularities of a large number of content instances, we can store the distinctive features of each more cheaply. Unsupervised learning methods such as autoencoders might be particularly well suited to this.

\begin{figure*}[htbp!]
\centering
\includegraphics[width=1.7\columnwidth]{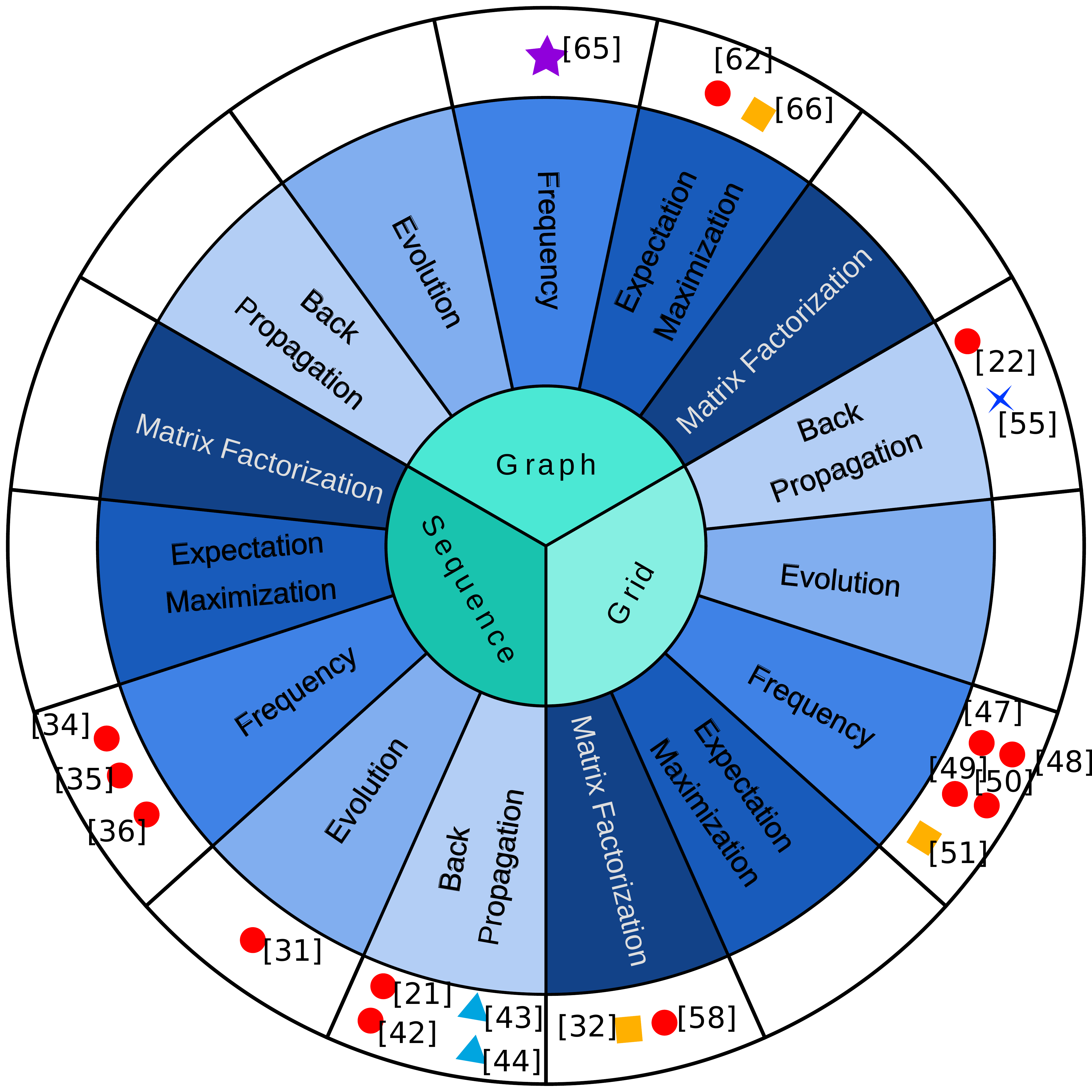}
\caption{Our taxonomization of PCGML techniques. We have two categorizations \textbf{(1)} the underlying data structure (graph, grid, or sequence) and \textbf{(2)} the training method (matrix factorization, expectation maximization, frequency counting, evolution, and back propagation).  Marks are colored for the specific type of content that was generated: red circles are platformer levels, orange squares are ``dungeons'', the dark blue x is real time strategy levels, light blue triangles are collectible game cards, and the purple star is interactive fiction. Citations for each are listed.   \label{fig:taxonomy}}
\end{figure*}
\section{Methods of PCGML}\label{sec:Methods}

We organize PCGML techniques using the following two dimensions:

\begin{itemize}[leftmargin=*]
\item \textbf{Data Representation} The underlying representation of the data used for training and generation. We consider three representations: \textit{Sequences}, \textit{Grids}, and \textit{Graphs}.  We note that it is possible for the same type of content to be represented in many different formats (e.g., platformer levels have been represented as all three representations), and that wildly different content can be represented in the same format (e.g., levels and \textit{Magic} cards as sequences) .
\item \textbf{Training Method} The machine learning technique utilized for training the model. %machine learning system.  
We consider five broad categories of training algorithms: \textit{Back Propagation}, \textit{Evolution}, \textit{Frequency Counting}, \textit{Expectation Maximization}, and \textit{Matrix Factorization}.  We note that it is both possible for the same underlying machine learned representation to be trained via different techniques and for two different techniques to utilize the same underlying class of training method (e.g., neural networks can be trained via back propagation \cite{isaksen2015discovering} \cite{summerville2016mariostring} \cite{summerville2016mtg} or evolution \cite{hoover:icccws15} and \textit{Expectation Maximization} can be used to train a Bayesian Network \cite{summerville2015samplinghyrule} or K-Means centroids \cite{guzdial2016learning}).
\end{itemize}

\noindent This organization has the benefits of highlighting commonalities across different techniques and game content. Figure \ref{fig:taxonomy} shows a graphical representation of the different approaches that have been attempted.  In the following sections we highlight different PCGML methodologies according to this taxonomy and discuss potential future work to address gaps in the coverage of the prior approaches. % that have been attempted.

\subsection{Sequences}

Sequences represent a natural format for contant that is experienced over time, such as textual content (\textit{Magic} cards) and game levels.  We note that the only game levels that has been handled as a sequence have come from the early \textit{Super Mario Bros.} games where the player can only traverse from left-to-right, meaning that there is a natural ordering of the two-dimensional space into a one-dimensional sequence.
\subsubsection{Frequency Counting}

Frequency counting refers to methods wherein the data is split and the frequencies of each type of atomic generative piece (e.g., tile for a tilemap based game) are found, determining the probabilities of generation. These need not simply be the raw frequencies, but are more likely the conditional probability of a piece given some state. Markov chains are a class of techniques that learn conditional probabilities of the next state in a sequence based on the current state.  This state can incorporate multiple previous states via the construction of $n$-grams. An $n$-gram model (or $n$-gram for short) is simply an $n$-dimensional array, where the probability of each state is determined by the $n$ states that precede it. 

\begin{figure}[tbp!]
\centering
\begin{subfigure}{\linewidth}
  \centering
  \includegraphics[width=.9\linewidth]{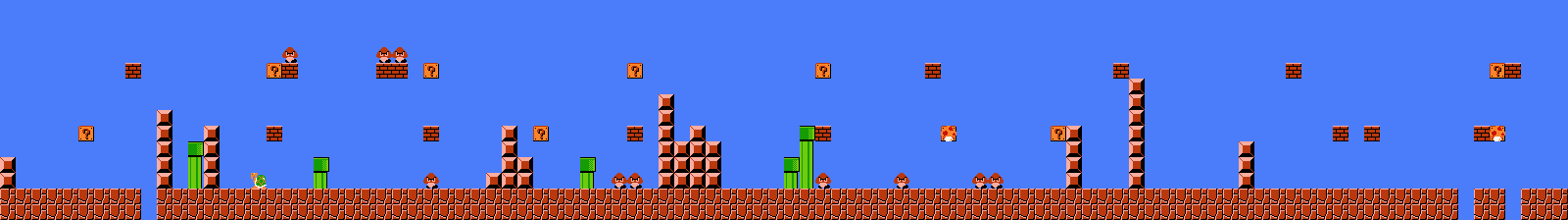}
  \caption{\centering $n=1$}
\end{subfigure}
\begin{subfigure}{\linewidth}
  \centering
  \includegraphics[width=.9\linewidth]{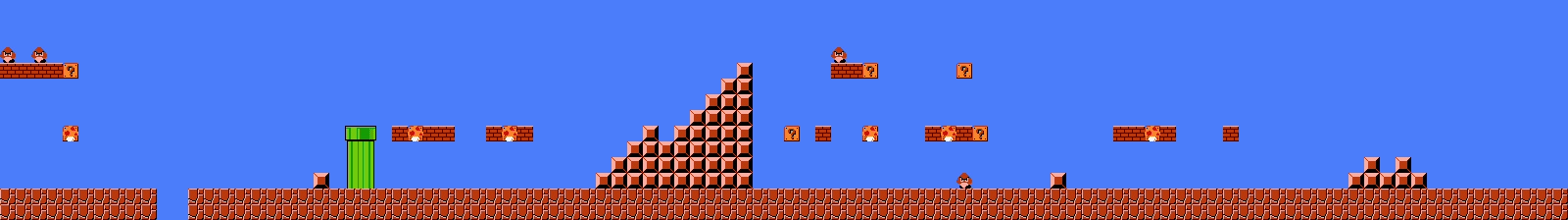}
  \caption{\centering $n=2$}
\end{subfigure}
\begin{subfigure}{\linewidth}
  \centering
  \includegraphics[width=.9\linewidth]{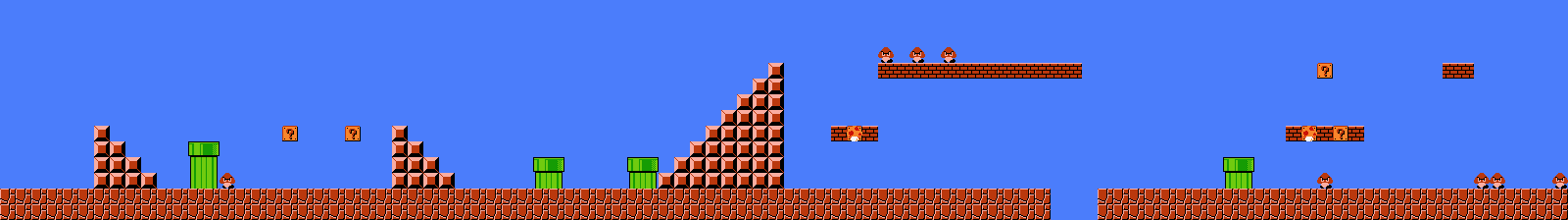}
  \caption{\centering $n=3$}
\end{subfigure}
\caption{Mario levels reconstructed by $n$-grams with $n$ set to 1, 2, and 3 respectively. Figures reproduced with permission from ~\cite{dahlskog2014linear}.}
\label{fig:ngram}
\end{figure}

Dahlskog et al. trained $n$-gram models on the levels of the original \textit{Super Mario Bros.} game, and used these models to generate new levels~\cite{dahlskog2014linear}. As $n$-gram models are fundamentally one-dimensional, these levels needed to be converted to strings in order for $n$-grams to be applicable. This was done through dividing the levels into vertical ``slices,'' where most slices recur many times throughout the level~\cite{dahlskog2013patterns}. This representational trick is dependent on there being a large amount of redundancy in the level design, something that is true in many games. Models were trained using various levels of $n$, and it was observed that while $n=0$ creates essentially random structures and $n=1$ creates barely playable levels, $n=2$ and $n=3$ create rather well-shaped levels. See Figure~\ref{fig:ngram} for examples of this. 

Summerville et al. ~\cite{summervillemcmcts} extended these models with the use of Monte Carlo Tree Search (MCTS) to guide generation.  Instead of solely relying on the learned conditional probabilities, they used the learned probabilities during roll-outs (generation of whole levels) that were then scored based on an objective function specified by a designer (e.g., allowing them to bias the generation towards more or less difficult levels).  The generated levels could still only come from observed configurations, but the utilization of MCTS meant that playability guarantees could be made and allowed for more designer control than just editing of the input corpus.

\begin{figure*}[tbp!]
\centering
\includegraphics[width=1.7\columnwidth]{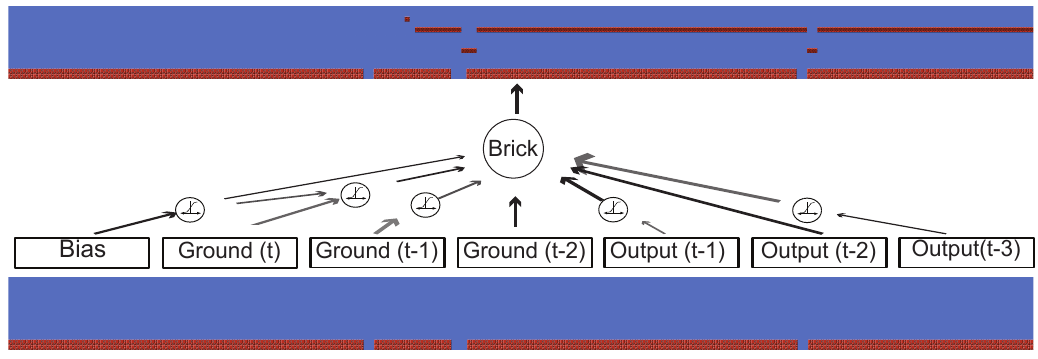}%{iccc_ws_ann.pdf}
\caption{Visualization of the NeuroEvolution approach, showing the input (bottom), an evolved network architecture, and an example output (top). Figures reproduced with permission from  \cite{hoover:icccws15}.\label{fig:ANN}}
%\caption{{\bf The ANN Representation}. At each column in a given level, an ANN inputs visual assets (i.e. musical voices) from at least one tile type of  \textit{Super Mario Bros.} and outputs a different type of visual asset. This example figure shows ground tiles being input to the ANN while the brick tile placements are output from predictions made through NeuroEvolution of Augmenting Topologies \cite{stanley:ec02}. To best capture the regularities in a level, each column is also provided information about the most recently encountered inputs (i.e. input values for the three previous columns). The inputs and outputs are then fed back into the ANN at each subsequent tick. Once trained, ANNs can potentially suggest reasonable placements for new human composed in-game tiles.  Figures reproduced with permission from  \cite{hoover:icccws15}.\label{fig:ANN}}
\end{figure*}

\subsubsection{Evolution}

%AMY: Evolutionary approaches rely on a genotype 
%AMY: Maybe on some level we can look at all genotypes as array of numbers but that description minimizes the importance of structures behind determining these arrays
%AMY (typically represented as an array of numbers) 
%AMY that represent a parameterization of the type of instance that is being optimized.  
%%AMY/JULIAN add in some good evo examples
%AMY While evolutionary approaches have been used to generate content, they are not under the purview of what we are considering here.  Instead, we are looking at evolutionary approaches that are used to generate generators where the objective function is derived from how well the generator is able to match the dataset.  

Evolutionary approaches parameterize solutions as genotypes that are then translated and evaluated in phenotypic or behavior space. This section focuses on evolutionary approaches to generate generators (rather than more general content) where the objective function is based on the generator's ability to match a dataset. 

% \subsection{Artificial Neural Networks}
% Recently neural networks have been used in the context of generating images \cite{mordvintsev2015inceptionism, gatys2015neural} and learning to play games \cite{mnih2015human, silver2016mastering, samothrakis2015neuroevolution} with great success. Unsurprisingly, neural networks are also being explored in PCG for level and map generation as well as cooperative trading card generation.

Hoover et al. \cite{hoover:icccws15} generate levels for \textit{Super Mario Bros.} by extending a representation called functional scaffolding for musical composition (FSMC) that was originally developed to compose music.  The original FSMC evolves musical voices to be played simultaneously with an original human-composed voice  \cite{hoover:cmj14} via  NeuroEvolution of Augmenting Topologies (NEAT) \cite{stanley:ec02}. % The original FSMC representation posits 1)~music can be represented as a function of time and  2)~musical voices in a given piece are functionally related \cite{hoover:cmj14}. Through a method for evolving Artificial Neural Networks (ANNs) called NeuroEvolution of Augmenting Topologies (NEAT) \cite{stanley:ec02}, additional musical voices are evolved to be played simultaneously with an original human-composed voice. 

To extend this musical metaphor and represent \textit{Super Mario Bros.} levels as functions of time, each level is broken down into a sequence of tile-width columns. Additional voices or types of tiles are then evolved with ANNs trained on two-thirds of the existing human-authored levels to predict the value of a tile-type at each column (as shown in Figure \ref{fig:ANN}). 
This approach combines a minimal amount of human-authored content with the output from previous iterations. The output of the network is added to the newly created level and fed back as input into the generation of new tile layouts. By acknowledging and iterating on the relationships between design pieces inherent in a human-produced level, this method can generate maps that both adhere to some important aspects of level design while deviating from others in novel ways.

Many evolutionary and evolutionary-like algorithms generate content through machine learning based fitness functions, but because generation happens through an author-defined search space they are not PCGML. Another interesting combination of search and machine learning for PCG is the DeLeNoX algorithm, which uses unsupervised learning to continuously reshape the search space and novelty search to search for content to explore the search space~\cite{liapis2013transforming}.

\subsubsection{Back Propagation}

Artificial Neural Networks are universal function approximators and have seen use in the field of PCGML as an approximator for a designer.  ANNs can be trained via evolutionary techniques as discussed in the previous section, but they are commonly trained via back propagation.  Back propagation refers to the propagation of errors through the ANN, with each weight in the ANN being changed proportional to the amount of error for which it is responsible.  Where the evolutionary approaches are only able to score the entire network, back propagation tunes parts of the network responsible for errors; however, this comes at the cost that all functions in the ANN must be differentiable (which evolutionary approaches do not require).  

Summerville and Mateas \cite{summerville2016mariostring} used Long Short-Term Memory Recurrent Neural Networks (LSTM RNNs) \cite{hochreiterLSTM} to generate levels learned from a tile representation of \textit{Super Mario Bros.} and \textit{Super Mario Bros. 2} (JP) \cite{supermariobros2} levels.  LSTMs are a variant of RNNs that represent the current state-of-the-art for sequence based tasks, able to hold information for 100's and 1000's of time steps, unlike the 5-6 of standard of RNNs. Summerville and Matteas used a tile representation
%like the one used by Snodgrass and Onta{\~n}{\'o}n \cite{snodgrass2014experiments}, but instead of 
representing levels as %2-D arrays they represent them as
a linear string with 3 different representations used for experimentation.  They also included simulated player path information in the input data, forcing the generator to generate exemplar paths in addition to the level geometry. Finally, they included information about how deep into the string the level geometry was, causing the generator to learn both level progression and when a level should end.  

Summerville et al. \cite{summerville2016playerTailored} extended this work to incorporate actual player paths extracted from 4 different YouTube playthroughs.  
The incorporation of a specific player's paths biased the generators to generate content suited to that player (e.g., the player that went out of their way to collect every coin and question-mark block had more coins and question-marks in their generated levels).

\begin{figure}[ptp!]
\centering
\includegraphics[width=0.4\textwidth]{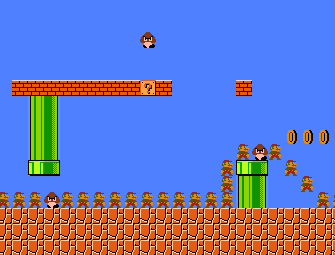}
\caption{Example output of the LSTM approach, including generated exemplar player path. Figure reproduced with permission from \cite{summerville2016mariostring}.}
\end{figure}

A rare example of game content generation that is not a level generator is \textit{Magic: The Gathering} \cite{mtg} card generation.  The first to use a machine learning approach for this is Morgan Milewicz with \textit{@RoboRosewater} \cite{ROBOROSEWATER} a twitter bot that uses LSTMs to generate cards.  Trained on the entirety of the corpus of Magic cards it generates cards, represented as a sequence of text fields (e.g., Cost, Name, Type, etc.).  A limitation of this representation and technique is that by treating cards as a sequence there is no way to condition generation of cards on fields that occur later in the sequence (e.g., Cost occurs near the end of the sequence and as such can not be used to condition the generation unless all previous fields are specified).

Building on the work of \textit{@RoboRosewater} is \textit{Mystical Tutor} by Summerville and Mateas \cite{summerville2016mtg}.  Using sequence-to-sequence learning \cite{seq2seq}, wherein an encoding LSTM encodes an  input sequence into a fixed length vector which is then decoded by a decoder LSTM, they trained on the corpus of  \textit{Magic: The Gathering}.  They corrupted the input sequences by replacing lines of text with a \texttt{MISSING} token and then tried to reproduce the original cards as the output sequence.  The corruption of the input along with the sequence-to-sequence architecture allows the generator to be conditioned on any piece of the card, addressing one of the limitations of \textit{@RoboRosewater}. This work shows multiple different paradigms for PCGML, showing how it can be used on different game content and as a co-creative design assistant, as in Hoover et al.'s \cite{hoover:icccws15} previously mentioned work.

\begin{figure}[tbp!]
\centering
\includegraphics[width=0.4\textwidth]{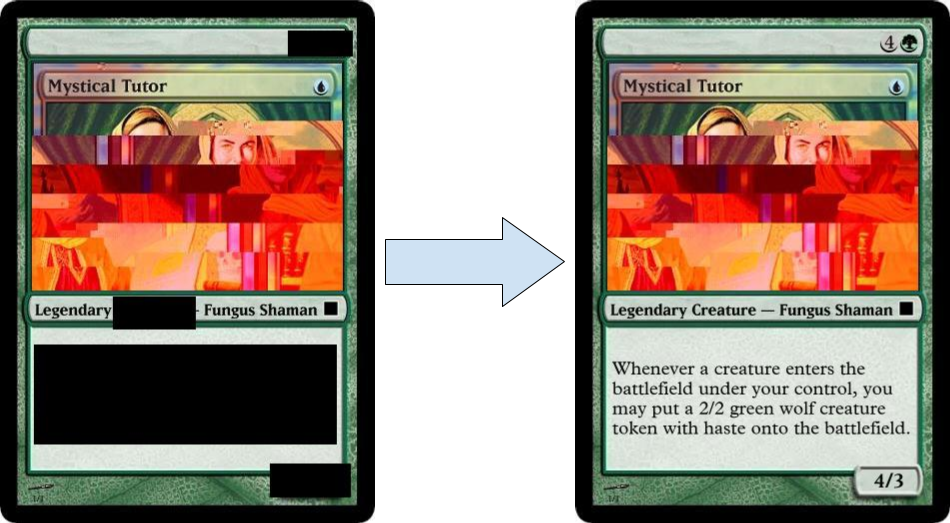}
\caption{Partial card specification and the output. Figure reproduced with permission from \cite{summerville2016mtg}.}
\end{figure}

\subsection{Grids}

Most game levels (particularly of the pre-3D era) can be thought of as two-dimensional grids.  Sometimes these representations are lossy (e.g. a non-tile entity is forcibly aligned to the grid, even if it could actually be at a non-tile position), but are generally a natural representation for many different kinds of levels (e.g. platformers, dungeons, real time strategy maps, etc.).  

\subsubsection{Frequency Counting}

An extension to the previously discussed one dimensional Markov chains are Multi-dimensional Markov Chains (MdMCs) \cite{ching2007multi}, wherein the state represents a surrounding neighborhood and not just a single linear dimension.  Snodgrass and Onta{\~n}{\'o}n  \cite{snodgrass2014experiments} present an approach to level generation using MdMCs. An MdMC differs from a standard Markov chain in that it allows for dependencies in multiple directions and from multiple states, whereas a standard Markov chain only allows for dependence on the previous state alone. In their work, Snodgrass and Onta{\~n}{\'o}n represent video game levels as 2-D arrays of tiles representing features in the levels. For example, in \textit{Super Mario Bros.} they use tiles representing the ground, enemies, and {\em ?}-blocks, etc. These tile types are used as the states in the MdMC. That is, the type of the next tile is dependent upon the types of surrounding tiles, given the network structure of the MdMC (i.e. the states that the current state's value depends on).

They train an MdMC by building a probability table according to the frequency of the tiles in training data, given the network structure of the MdMC, the set of training levels, and the set of tile types. A new level is then sampled one tile at a time by probabilistically choosing the next tile based upon the types of the previous tiles and the learned probability table.

\begin{figure}[tbp!]
\centering
\includegraphics[width=0.465\textwidth]{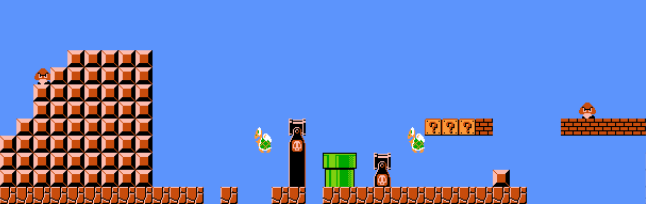}\\
\vspace{2mm}
\includegraphics[height=0.2\textheight]{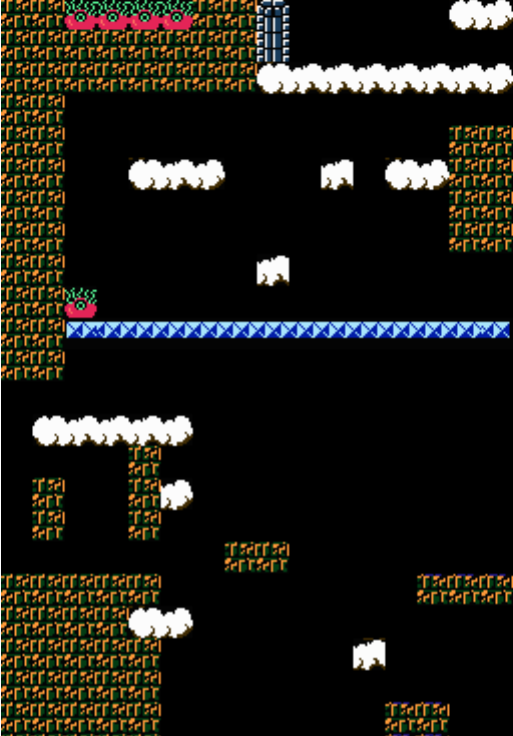}
\includegraphics[height=0.2\textheight]{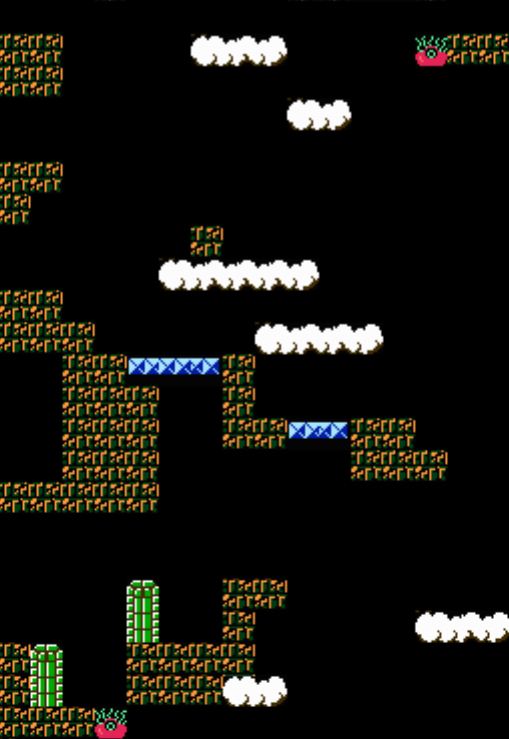}
\caption{Sections from a \textit{Super Mario Bros.} level (top) and \textit{Kid Icarus} level section both generated using the constrained MdMC approach (bottom-left), and using an MRF approach (bottom right). Figures reproduced with permission from \cite{snodgrassconstrained,snodgrass2016learning}.}
\label{fig:mdmc}
\end{figure}

In addition to their standard MdMC approach, Snodgrass and Onta{\~n}{\'o}n have explored hierarchical \cite{snodgrass2015hierarchical} and constrained \cite{snodgrassconstrained} extensions to MdMCs in order to capture higher level structures and ensure usability of the sampled levels, respectively. They have also developed a Markov random field approach (MRF) \cite{snodgrass2016learning} that performed better than the standard MdMC model in \textit{Kid Icarus}, a domain where platform placement is pivotal to playability. Figure~\ref{fig:mdmc} shows a section of a \textit{Super Mario Bros.} level (top) and a section of a \textit{Kid Icarus} level (bottom-left) sampled using a constrained MdMC approach, and a section of \textit{Kid Icarus} level (bottom-right) sampled using the MRF approach.

A recent approach by Gumin \cite{Gumin2016} is loosely inspired by quantum mechanics and uses a ``superposition'' of tiles to generate images and levels from a representative example tile set. This approach is a variant of MRF, except instead of solely sampling, samples are chosen via ``collapsing of the wave function'' (i.e. probabilistically choosing a tile and propagating constraints that that choice enforces).  This in turn can propagate other changes and either deterministically chooses tiles that no longer have any other possible choices or reduces the possible set of other tiles.  The probabilities and configurations are determined by finding each $N\times N$ window in the input, and the number of times that window occurs. This approach was initially explored for bitmap generation, but has since been expanded for use with 3-D tile sets as well as for level generation \cite{qud,skate}. The source code and examples of the bitmap project can be found online \cite{Gumin2016}.

\subsubsection{Back Propagation}

\begin{figure}[tbp!]
\centering
\begin{subfigure}{0.32\linewidth}
  \centering
  \fbox{\includegraphics[width=.5\linewidth]{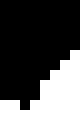}}
  \caption{\centering Original}
  \label{fig:repair:1}
\end{subfigure}%
\begin{subfigure}{0.32\linewidth}
  \centering
  \fbox{\includegraphics[width=.5\linewidth]{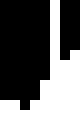}}
  \caption{\centering Unplayable}
  \label{fig:repair:2}
\end{subfigure}
\begin{subfigure}{0.32\linewidth}
  \centering
  \fbox{\includegraphics[width=.5\linewidth]{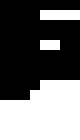}}
  \caption{\centering Repaired}
  \label{fig:repair:3}
\end{subfigure}
\caption{The original window is overwritten with a wall making the game unplayable.  The autoencoder repairs the window to make it playable, although it chooses a different solution to the problem. Figures reproduced with permission from \cite{jain2016autoencoders}.}
\label{fig:repair}
\end{figure}

In \cite{jain2016autoencoders} Jain et al. show how autoencoders \cite{vincent2008extracting} may be trained to reproduce levels from the original \textit{Super Mario Bros.} game.
The autoencoders are trained on series of vertical level windows and compress the typical features of Mario levels into a more general representation. They experimented with the width of the level windows and found that four tiles seems to work best for Mario levels.
They proceeded to use these networks to discriminate generated levels from original levels, and to generate new levels as transformation from noise.
They also demonstrated how a trained autoencoder may be used to repair unplayable levels, changing tile-level features to make levels playable, by inputting a broken/unplayable level to the autoencoder and receiving a repaired one as output (see Figure~\ref{fig:repair}).

\begin{figure}[tbp!] 
\centering
%\begin{subfigure}{0.9\linewidth}
%  \centering
%  \includegraphics[width=\linewidth]{starcraftnets/1B}
%  \caption{\centering A \textit{ StarCraft II} heightmap}
%  \label{fig:starcraftnets:heightmap}
%\end{subfigure}

%\vspace{\baselineskip}

\begin{subfigure}{0.9\linewidth}
  \centering
  \includegraphics[width=\linewidth]{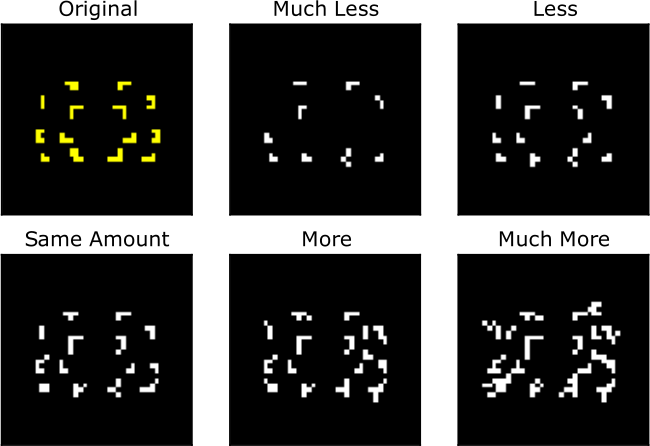}
  %\caption{\centering Varying resource amounts.}
  \label{fig:starcraftnets:lessmore}
\end{subfigure}
%\caption{A heightmap (a) and generated resource locations (b) for \textit{ StarCraft II}}
\caption{Varying resource amounts when generating resource locations for \textit{StarCraft II}.  Figures reproduced with permission from \cite{lee2016predicting}.}
\label{fig:starcraftnets}
\end{figure}

Lee et al.~\cite{lee2016predicting} use convolutional neural networks to predict resource locations in maps for \emph{StarCraft II}~\cite{starcraft2}.
Resource locations are sparsely represented in \textit{StarCraft II} maps, but are decisive in shaping gameplay. They are usually placed to match the topology of maps.
Exploiting this fact, they transform \textit{StarCraft II} maps into heightmaps, downsample them, and train deep neural networks to predict resource locations.
The neural networks are shown to perform well in some cases, but struggled in other cases, 
most likely due to overfitting the small training dataset.
By adding a number of postprocessing steps to the network, the authors created a tool that allows map designers to vary the amount of resource locations in an automatically decorated map, varying the frequency of mineral placements, shown in Figure~\ref{fig:starcraftnets}.

\subsubsection{Matrix Factorization}

Some approaches to level generation find latent level features in high-dimensional data through matrix factorization, which infers features by compressing existing data into a series of smaller matrices. While often generators create levels with a limited expressive range \cite{horn2014comparative}, Shaker and Abou-Zleikha \cite{shaker2014alone} create more expressive \textit{Super Mario Bros.} levels by first generating thousands with five known, non-ML-based generators (i.e.\ Notch, Parameterized, Grammatical Evolution, Launchpad, and Hopper). These levels are then compressed into vectors indicating the content type at each column and transformed into $T$ matrices for each type of content: Platforms, Hills, Gaps, Items, and Enemies. Through a multiplicative update algorithm \cite{lee1999learning} for non-negative matrix factorization (NNMF), these levels are factored into $T$ approximate ``part matrices'' which represent level patterns and $T$ coefficient matrices corresponding to weights for each pattern. The part matrices can be examined to see what types of patterns appear globally and uniquely in different generators, and multiplying these part matrices by novel coefficient vectors can be used to generate new levels.  This allows the NNMF method to explore far outside the expressive range of the original generators.

\begin{figure*}[tbp!]
\centering
\includegraphics[width=2\columnwidth]{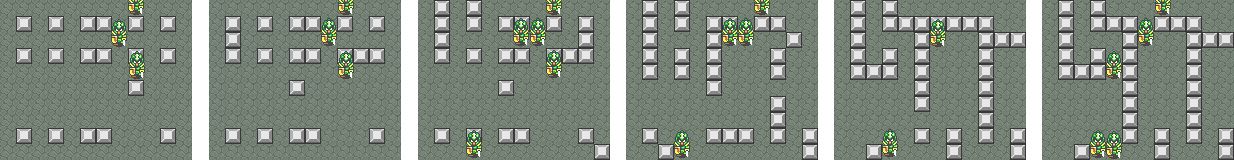}
\caption{Example of interpolation between two Zelda rooms (the leftmost and rightmost rooms). Figure reproduced with permission from \cite{summerville2015samplinghyrule}.}
\label{fig:pca_zelda}
\end{figure*}

While approaches to generating levels typically focus on platformers (e.g. \textit{Super Mario Bros.}), Summerville et al. \cite{summerville2015samplinghyrule} generate levels for  \textit{The Legend of Zelda} \cite{zelda} series.  Their approach relies on segmenting data hierarchically by first generating the high-level topological structure of a dungeon (discussed in Section \ref{sec:graph_em}) and then the rooms represented as a grid of tiles via Principal Component Analysis (PCA). The PCA algorithm finds a compressed representation of the original 2-D room arrays by taking the eigenvectors of the original high dimensional space and retaining only the most informative (inspired by EigenFaces\cite{eigenfaces}). These compressed representations are represented as weight vectors that can be interpolated between to generate new room content as seen in Figure \ref{fig:pca_zelda}.

\subsection{Graphs}\label{sec:graph_em}

Graphs are the most general data representation considered, with the previous data representations being easily representable as graphs.  However, this generality comes at a cost, which is the lack of well-defined structural properties (e.g. in a grid, above and below are implicitly encoded in the structure).  

\subsubsection{Expectation Maximization}

\begin{figure}[btp!]
\centering
\includegraphics[width=0.5\textwidth]{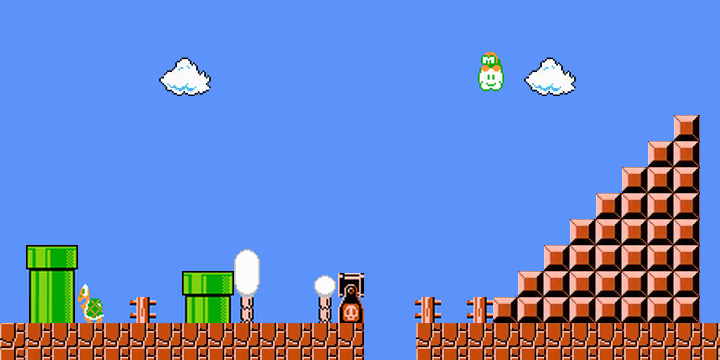}
\caption{Level section from the clustering approach. Figure reproduced with permission from \cite{guzdial2016learning}.}
\label{fig:clustering}
\end{figure}

Expectation Maximization (EM) is an iterative algorithm that seeks to find the Maximum Likelihood Estimate (MLE) or Maximum A Posteriori (MAP) estimate for the parameters of a model.  This is done via a two step process, first an Expectation (E) step wherein the current model's likelihood is computed given the training data and a Maximization (M) step where the parameters are changed to maximize the previously calculated likelihood (which are used in the next E step).  EM is a general training method for any model that is able to provide a likelihood of the model given training data.

K-means is a clustering method where an a priori defined number of clusters are computed based on the means of the data.  Each E step determines which data points belong to which cluster, resulting in new means being calculated for each cluster in the M step. Guzdial and Riedl~\cite{guzdial2016game} used a hierarchical clustering approach using K-means clustering with automatic K estimation to train their model. They utilized gameplay video of individuals playing through \textit{Super Mario Bros.} to generate new levels. They accomplished this by parsing each \textit{Super Mario Bros.} gameplay video frame-by-frame with OpenCV \cite{Pulli:2012:RCV:2184319.2184337} and a fan-authored spritesheet, a collection of each image that appears in the game. Individual parsed frames could then combine to form chunks of level geometry, which served as the input to the model construction process. In total Guzdial and Riedl made use of nine gameplay videos for their work with \textit{Super Mario Bros.}, roughly 4 hours of gameplay in total.

Guzdial and Riedl's model structure was adapted from \cite{kalogerakis:2012:SIG}, a graph structure meant to encode styles of shapes and their probabilistic relationships. The shapes in this case refer to collections of identical sprites tiled over space in different configurations. For further details please see \cite{guzdial2016game}, but it can be understood as a learned shape grammar, identifying individual shapes and probabilistic rules on how to combine them.  First the chunks of level geometry were clustered to derive styles of level chunks, then the shapes within that chunk were clustered again to derive styles of shapes, and lastly the styles of shapes were clustered to determine how they could be combined to form novel chunks of level. After this point generating a new level requires generating novel level chunks in sequences derived from the gameplay videos.  An example screen of generated content using this approach is shown in Figure~\ref{fig:clustering}.

\begin{figure}[tbp!]
\centering
\includegraphics[width=0.5\textwidth]{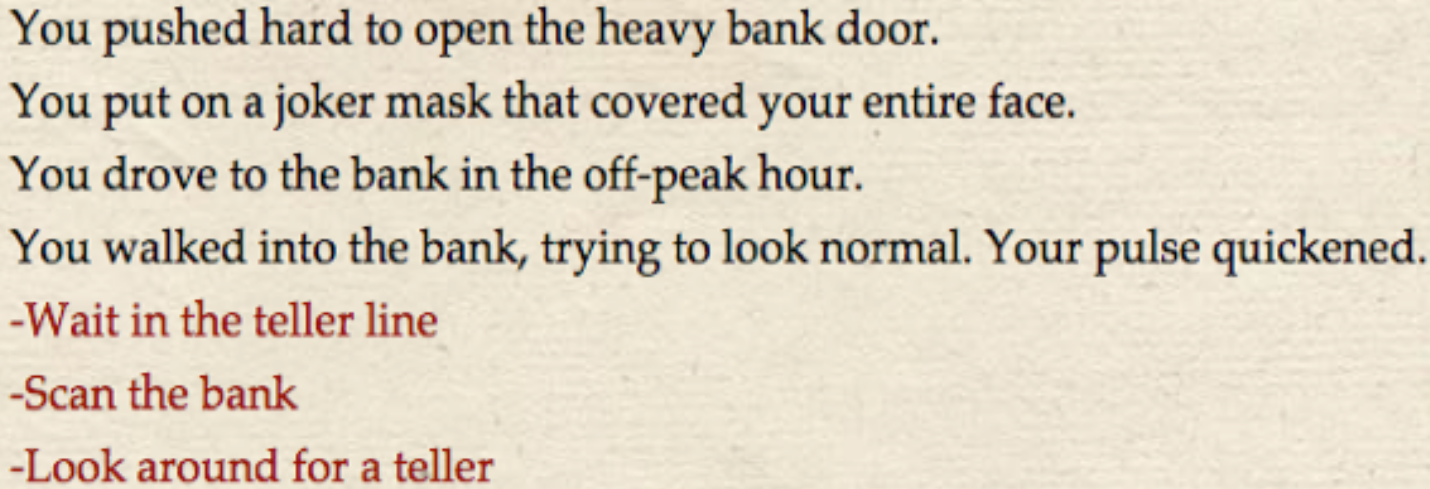}%{exampleScheherazade-IF.pdf}
\caption{Example of Scheherazade-IF gameplay. Figure reproduced with permission from \cite{guzdial2015crowdsourcing}.}
\end{figure}

As previously discussed, Summerville et al. \cite{summerville2015learning} generated dungeons using two different data representations, with the high level topological structure defined as a graph with rooms as nodes and linkages (e.g., doors, bombable walls, teleportation portals) as edges.   The room-to-room structure of a dungeon is learned along with high level parameters such as dungeon size and length of the optimal player path using a Bayes Net \cite{wright1921correlation}, a graphical structure of the joint probability distribution.  Bayes Nets are trainable via a number of different techniques (e.g., Gibbs Sampling, Variational Message Passing), but all are based broadly around maximizing the likelihood of a model given the data.  After training the Bayes Net, a designer can ``observe'' specific parameters of the model such as how many rooms it should have, how many rooms the player would need to traverse to complete the level, etc.  and then sample the rest room by room.  

\subsubsection{Frequency Counting}

PCGML has focused on graphical games, and particularly the levels of graphical games. However, there exists some work in the field of generating interactive fiction, text-based games like choose your own adventure stories. Guzdial et al. adapted Scheherazade \cite{li2015learning}, a system for automatically learning to generate stories, into Scheherazade-IF \cite{guzdial2015crowdsourcing}, which can derive entire interactive fiction games from a dataset of stories.

Both Scheherazades rely on exemplar stories crowdsourced from Amazon Mechanical Turk (AMT). The reliance on crowdsourcing rather than finding stories ``in the wild'' allows Scheherazade to collect linguistically simple stories and stories related to a particular genre or setting (e.g., a bank robbery, a movie date, etc). 

Scheherazade-IF's structures its model as plot graphs \cite{weyhrauch1997guiding}, directed graphs with events as vertices and sequentiality information encoded as edges. The system derives events from the individual sentences of the story, learning what events can occur via an adaption of the OPTICS clustering algorithm \cite{ankerst1999optics}.  The clusters found by OPTICS are found by only allowing points in a cluster if it keeps the density of the cluster the same.  Unlike K-means, OPTICS is able to find clusters of arbitrary shape, assuming the shapes are of consistent density. The density is determined by looking in a region around each point and counting the number of other points. The ordering of these primitive events can then be derived from the exemplar stories, which enables the construction of a plot graph for each type of story (e.g., bank robbery). Scheherazade-IF uses these learned plot graphs as the basis for an interactive narrative experience, allowing players to take the role of any of the characters in the plot graph (robber or bank teller in bank robbery) and make choices based on the possible sequences of learned events.
\subsection{Discussion of Approaches}

Broadly, we see that platformer level generation is the most common target of generation and is represented by all types of training and all types of representation.  The graph representation is the most general in form \cite{guzdial2016game}, with no notion of level shape or size -- only the relative positions of entities.  The sequence and grid based approaches all explicitly encode some aspect of the level shape, either by fixing the height at generation time \cite{dahlskog2014linear,hoover:icccws15,summervillemcmcts,summerville2016mariostring,summerville2016playerTailored} or at run time \cite{snodgrass2014experiments,snodgrass2015hierarchical,snodgrass2016approach,snodgrass2016learning,snodgrassconstrained} --- while the matrix factorization approach requires fixing both height and width \cite{shaker2014alone}.  Similarly, the convolutional grid approach \cite{isaksen2015discovering} for RTS map generation requires a fixed map size.

It is hard to compare and contrast the different approaches' performance as there is no agreed upon test for generator ``goodness'' (i.e. Playability of levels? Capturing of original levels' style? etc.) and the different approaches do not list the time it takes to train or generate a level.  However, a few general statements can be made.  The LSTM approaches \cite{summerville2016mariostring,summerville2016playerTailored} will nearly certainly take more time to train and generate than the MdMC approaches.  Both generate one tile at a time, but the LSTM approach inherently requires more computation to generate.  Furthermore, the MdMC can be trained in a single pass over the levels, but it is highly unlikely that training of the LSTM can be stopped after a single training epoch.  We note that the Matrix Factorization approaches have potentially the worst memory usage, as all levels must be held in memory at once.  In practice, given the relatively small datasets, this is unlikely a concern, but could pose an issue for situations that call for much larger or many more levels.  The back-propagation neural network approaches assume a fixed architecture and as such are the only training size independent approach. In practice, the generator size for the other approaches (MdMC, latent style-graph, evolved neural network, matrix factorized) are likely to be smaller than the back-propagated neural network, but there are no guarantees.  A similar concern for the latent style-graph approach is that generation time is dependent on the complexity of the training data (more dense, less regular levels will have more latent nodes and subsequently more relative position edges) unlike the other approaches which perform the same generation act at each step of generation.

\subsection{Unexplored Approaches}

From Figure \ref{fig:taxonomy} we see that only slightly over half of the possible combinations have been touched on.  Notably, evolutionary approaches have only been used to train ANNs for sequences, but given the generality of evolutionary approaches it seems possible for generators for both graph and grid approaches (given an appropriate objective function).  EM approaches are commonly used for training Hidden Markov Models (HMMs) \cite{hmm} where the states are hidden from observation and only the output of a hidden state is observed.  HMMs seem like a logical extension to the existing Markov chain work, as they allow for higher order structures to be learned instead of just column-to-column transitions.  Extending Markov Random Fields similarly, Conditional Random Fields (CRF) \cite{crf} act as if the connected grid is based on  an unobserved, latent state and have been used in image generation \cite{crf_imagegen}.  

While it may seem that matrix factorization techniques would be ill-suited to sequences and graphs, those are actually fruitful areas of research.  Global Vectors (GloVe) for word representation \cite{glove} utilize matrix factorization to embed categorical vocabularies into lower-dimensional real-valued vectors.  These vectors are then used for analogical reasoning \cite{emoji2vec} or for generation \cite{allison_parrish}, but have yet to see use in the generation of game content.  Graphs can be represented as adjacency matrices, where the rows and columns are the nodes and each entry in the matrix represents whether nodes are connected (1) or not (0).  This enables the field of spectral graph theory, and it seems possible that matrix factorization techniques could learn important properties of graphs represented this way.% from such a representation.  

Finally, until recently graphs have been an unexplored data structure for back propagation based neural networks.  Recent work from Kipf and Welling ~\cite{kipf2016semi} on graph convolutional networks have only been used for classification purposes, but given the lag between pixel convolutional networks being used for classification (1998) \cite{lecun1998mnist} and for generation (2015) \cite{dosovitskiy2015learning} this seems more a matter of time than a matter of possibility.

\section{Open Problems and Outlook}
\label{sec:open-problems}
In this section, we describe some of the major issues facing PCGML and open areas for active research.  Because games as a ML research domain are considerably different from other ML domains, such as vision and audio learning, the problems that arise are commensurately different.
In particular, game datasets are typically much smaller than other domains' datasets, they are dynamic systems that rely on interaction, and the availability of high-quality clean data is limited.  In the following sections we discuss the issue of limited data as well as underexplored applications of PCGML for games including style transfer, parameter tuning, and potential PCGML game mechanics.

\subsection{Ensuring Solvability and Playability}

A key aspect of procedural generation for games is the generation of playable artifacts.  So far, most of this work has relied on designer specified constraints on the generated works, either via constraints after generation \cite{snodgrassconstrained} or applying constraints at each step of sampling \cite{summervillemcmcts}.  While there have been approaches that utilize player paths (either synthetic \cite{summerville2016mariostring} or actual \cite{summerville2016playerTailored}), these do not ensure playability but instead bias the generators to create playable levels.  Future work is required to train a generator to only generate playable levels, but we envision this could be achieved via penalizing the generator at training time, as in generative adversarial networks.

\subsection{Data Sources and Representations}
\label{sec:data-sources}
In this section we investigate how training data, varying in source, quality, and representation, can impact PCGML work. A commonly held view in the machine learning community is that more data is better.  Halevy, et al. \cite{UnreasonableEffectivenessOfData} make the case that having access to more data, not better algorithms or better data, is the key difference between why some problems are able to be solved in an effective manner with machine learning while others are not (e.g., language translation is a much harder problem than document classification, but more data exists for the translation task). For some types of machine learning a surfeit of data exists enabling certain techniques, such as images for generative adversarial networks \cite{SomeGANPaper} or text for Long Short-Term Memory recurrent neural networks (LSTMs) \cite{Karpathy_charRNN}; however, video games do not have this luxury.  While many games exist, they typically share no common data structures. Additionally, even if they do they typically do not share semantics (e.g., sprites in the NES share similarity in how they are displayed but sprite \texttt{0x01} is unlikely to have similar semantics between \textit{Super Mario Bros.} \cite{supermariobros1} and \textit{Metroid} \cite{metroid}, despite the fact that both are platformers).  As such, data sources are often limited to within a game series \cite{summerville2015samplinghyrule,summerville2016mariostring} and more commonly within individual games \cite{dahlskog2014linear,snodgrass2015hierarchical,guzdial2016game}.  This in turn means that data scarcity is likely to plague PCGML for the foreseeable future, as there is a small, finite amount of official artifacts (e.g., maps) for most games.  However, for many games there are vibrant fan communities producing hundreds, thousands, and even millions of artifacts, such as maps \cite{ArticleAboutMarioMaker}, which could conceivably provide much larger data sources. Within the text of this paper we discuss approaches to artifically increase the size of datasets through corruption of training data in Magic cards, including multiple player paths for platformer levels, and gathering training data from a secondary source such as video.

Recently, Summerville et al. created the video game
 level corpus (VGLC) \cite{summerville2016vglc}, a collection of video game levels represented in several easily parseable formats.  The corpus contains 428 levels from 12 games in 3 different file formats.  This marks an initial step in creating a shared data source that can be used for commonality across PCGML research.  That said, the repository is limited in scale (12 games and 428 levels is small for standard machine learning) and the representations are lossy (e.g., in Mario, levels only contain a single \textit{Enemy} type mapping for all Goombas, Koopa Troopas, etc.), so additional effort is needed to increase the scope and fidelity of the representations.

Assuming bountiful data existed for video games, bigger questions abound: 1) How should the data be represented? 2) What is a training instance?  For certain classes of PCGML, such as image generation, obvious data structures exist (e.g., RGB matrices), but there are no common structures for games.  Games are a combination of some amount of content (levels, images, 3-D models, text, boards, etc.) and the rules that govern them (What happens when input~$x$ is held down in state~$y$?  What happens if $z$ is clicked on?  Can player~$\alpha$ move piece~$p$? etc.).  Furthermore, the content is often dynamic during playtime, so even if the content were to capture with perfect fidelity (e.g., a Goomba starts at tile $X,Y$) how it will affect gameplay requires play (e.g., by the time the player can see the Goomba it has moved to tile $X',Y'$). Thus, issues of representation serve as a second major challenge, after access to sufficient quantities of quality data.

\subsection{Learning from Small Datasets}
As previously mentioned, it is a commonly held tenet that more data is better; however, games are likely to always be data-constrained.  The English Gigaword corpus \cite{gigaword} which has been used in over 450 papers at the time of writing is approximately 27 GB.  The entire library of games for the Nintendo Entertainment System is approximately 237 MB, over 2 orders of magnitude smaller.  The most common genre, platformers, makes up approximately $14\%$ of the NES library (and is the most common genre for PCG research) which is roughly another order of magnitude smaller.  For specific series it is even worse, with the largest series (\textit{Mega Man}~\cite{megaman}) making up $~0.8\%$ of the NES library, for 4 orders of magnitude smaller than a standard language corpus.  Standard image corpora are even larger than the Gigaword corpus, such as the ImageNet corpus \cite{imagenet} at 155GB.  All told, work in this area is always going to be constrained by the amount of data, even if more work is put into creating larger, better corpora.

While most machine learning is focused on using large corpora, there is work focused on One-Shot Learning/Generalization~\cite{feifeiOneShot}.  Humans have the ability to generalize very quickly to other objects of the same type after being shown a single object (or very small set, e.g., $n < 5$)  and One-Shot Learning is similarly concerned with learning from a very small dataset.  Most of the work has been focused on image classification, but games are a natural testbed for such techniques due to the paucity of data. 

\subsection{Learning on Different Levels of Abstraction}
As described above, one key challenge that sets PCGML for games apart from PCGML for domains such as images or sound is the fact that games are multi-modal dynamic systems.
This means that content generated will have interactions: generated rules operate on generated levels which in turn change the consequences of those rules and so on.
A useful high-level formal model for understanding games as dynamic systems that create experiences is the Mechanics, Dynamics, Aesthetics (MDA) framework by Hunicke et al.~\cite{hunicke2004mda}.
``Mechanics describes the particular components of the game, at the level of data representation and algorithms. Dynamics describes the run-time behavior of the mechanics acting on player inputs and each other's outputs over time. Aesthetics describes the desirable emotional responses evoked in the player, when she interacts with the game system.''
If the end goal for PCGML for games is to generate content or even complete games with good results at the aesthetic level, the problem is inherently a hierarchical one. 
It is possible that learning from data will need to happen at all three levels simultaneously in order to successfully generate content for games.

\subsection{Datasets and Benchmarks}
Procedural content generation, and specifically PCGML is a developing field of research. As such, there are not many publicly available datasets and no widely used standardized evaluation benchmarks.

Publicly available, large datasets are important, as they will reduce the barrier for entry into PCGML, which will allow the community to grow more quickly. As previously mentioned, the VGLC \cite{summerville2016vglc} is an important step in creating widely available datasets for PCGML. However, it currently only provides level data. Other data sets exist for object models\footnote{opengameart.org} and gameplay mechanics\footnote{http://www.squidi.net/three/index.php}, but they have not been utilized by the PCGML community. By bringing these data sets to the attention of researchers, we hope to promote the exploration and development of approaches for generating those types of content. Additionally, creating, improving, and growing both new and existing data sets is necessary for refining and expanding the field.

Widespread benchmarks allow for the comparison of various techniques in a standardized way. Previously, competitions have been used to compare various level generation approaches \cite{togelius2013mario,khalifaGVG}. However, many of the competitors in the \textit{Super Mario Bros.} competition included hard coded rules, and the GVG-AI competition provides descriptions of the games, but not the training data necessary for machine learning approaches.

In addition to competitions, there has been some work in developing metric-based benchmarks for comparing techniques. Horn et al. \cite{horn2014comparative} proposed a benchmark for \textit{Super Mario Bros.} level generators in a large comparative study of multiple generators, which provided several evaluation metrics that are commonly used now, but only for level generators for platforming games. In recent years, Canossa and Smith \cite{canossa2015towards} presented several general evaluation metrics for procedurally generated levels, which could allow for cross-technique comparisons. Again, however, these only apply to level generation and not other generatable content. Other machine learning domains, such as image processing, have benchmarks in place \cite{lecun1998mnist} for testing new techniques and comparing various techniques against each other. For the field of PCGML to grow, we need to be able to compare approaches in a meaningful way. 

\subsection{Style Transfer}

Style and concept transfer is the idea that information learned from one domain can enhance or supplement knowledge in another domain. Style transfer has most notably been explored for image modeling and generation \cite{bruckner2007style,hertzmann2001image}. For example, in recent work Gatys et al. \cite{gatys2015neural} used neural networks to transfer the style of an artist onto different images. Additionally, Deep Dream \cite{mordvintsev2015inceptionism} trains neural networks on images, and then generates new images that excite particular layers or nodes of the network. This approach could be adapted to learn from game content (e.g., levels, play data, etc.) and generate content that excites layers associated with different elements from the training data. More traditional forms of blending domain knowledge have been applied sparingly to games, such as game creature blending \cite{ribeiro2003model} and interactive fiction blending \cite{permar2013conceptual}. However, until very recently no one has applied this machine learning-oriented style transfer to procedural content generation. 

Recently Snodgrass and Onta{\~n}{\'o}n \cite{snodgrass2016approach} explored domain transfer in level generation across platforming games, and Guzdial et al. \cite{guzdial2016learning} used concept blending to meld different level designs from \textit{Super Mario Bros.} together (e.g., castle, underwater, and overworld levels). These approaches transfer and blend level styles, but do not attempt to address the game mechanics explicitly; both approaches ensure playable levels, but do not attempt transfer or blending between different mechanics. Gow and Corneli \cite{gow2015towards} proposed a framework for blending two games together into a new game, including mechanics, but no implementation yet exists.

The above approaches are important first steps, but style transfer needs to be explored more fully in PCG. Specifically, approaches to transferring more than just aesthetics are needed, including game mechanic transfer. The above approaches only apply transfer to game levels, but there are many other areas where style transfer can be applied (e.g., character models, stories, and quests).

\subsection{Exposing and Exploring the Generative Space}
One avenue that has yet to be deeply explored in PCGML is opening the systems up to allow for designer input.  Most systems are trained on a source of data and the editorial control over what data is fed in is the main interaction for a designer, but it is easy to imagine that a designer would want to control things such as difficulty, complexity, or theming (e.g., the difference between an above ground level and below ground level in \textit{Super Mario Bros.}).  The generative approach of Summerville et al. \cite{summerville2015samplinghyrule} allows a designer to set some high-level design parameters (number of rooms in a dungeon, length of optimal player path), but this approach only works if all design factors are known at training time and a Bayes Net is a suitable technique. Similarly, Snodgrass and Onta{\~n}{\'o}n \cite{snodgrassconstrained} allow the user to define constraints (e.g., number of enemies, distance of the longest gap, etc.), but require hand crafted constraint checkers for each constraint used.%, and for the domain to be able to be modeled with multi-dimensional Markov chains.
%Sam: Added a mention of our constraint-based paper as it seems relevant

The nature of Generative Adversarial Networks has allowed for the discovery of latent factors that hold specific semantics (e.g., subtracting the latent space image of blank-faced person from a smiling one is the smile vector) \cite{SmileVector}.  Interpolating and extrapolating along these latent dimensions allows a user to generate content while freely tuning the parameters they see fit.  We envision that this type of approach could lead to similar results in PCG (e.g., subtract Mario 1-1 from Mario 1-2 for the underground dimension, subtract Mario 1-1 from Mario 8-3 for the difficulty dimension).  Furthermore, if multiple games are used as training input, it is theoretically possible that one could interpolate between two games (e.g., find the half-way point between Mario and Sonic) or find other more esoteric combinations ($Contra - Mario + Zelda = ???$).

\subsection{Using PCGML as a Game Mechanic}
Most current work in PCGML focuses on replicating designed content to provide the player infinite or novel variations on gameplay, informed by past examples.
Another possibility is to use PCGML as the main mechanic of a game, presenting the PCGML system as an adversary or toy for the player to engage with.
Designs could include enticing the player to, for example, generate content that is significantly similar to or different from the corpus the system was trained on, identify content examples that are outliers or typical examples of the system, or train PCGML systems to generate examples that possess certain qualities or fulfill certain objective functions, teaching the player to operate a model by feeding it examples that shape its output in one direction or the other. This would allow a PCGML system to take on several design pattern roles, including AI as Role-Model, Trainee, Editable, Guided, Co-Creator, Adversary, and Spectacle~\cite{treanor2015ai}.\\
\textbf{Role-model:} a PCGML system replicates content generated by players of various levels of skill or generates content suitable for players of certain skill levels. New players are trained by having them replicate the content or by playing the generated content in a form of generative tutorial.\\
\textbf{Trainee:} the player trains a PCGML system to generate a piece of necessary content (e.g., part of a puzzle or level geometry).\\
\textbf{Editable:}
rather than training the AI to generate the missing puzzle piece via examples, the player changes the internal model's values until acceptable content is generated.\\
\textbf{Guided:} the player corrects PCG system output in order to fulfill increasingly difficult requirements. The AI, in turn, learns from the player's corrections, following the player's guidance.\\
\textbf{Co-creator:} the player and a PCGML system take turns in creating content, moving toward some external requirement. The PCGML system learns from the player's examples.\\
\textbf{Adversary:} the player produces content that the PCGML system must replicate by generation to survive or vice versa in a ``call and response'' battle.\\
\textbf{Spectacle:} the PCGML system is trained to replicate patterns that are sensorically impressive or cognitively interesting.

\section{Conclusion}
\label{sec:conclusion}
In this survey paper, we give an overview of an emerging machine learning approach to Procedural Content Generation, including describing and contrasting the existing examples of work taking this approach and outlining a number of challenges and opportunities for future research. We intend the paper to play a similar role as the Search-Based Procedural Content Generation paper~\cite{togelius2011search}, which pointed out existing research as well as work that was yet to be done. Much research that was proposed in that paper was subsequently carried out by various authors. 
%Removed table
There is much work left to do. The vast majority of work has so far concerned two-dimensional levels, in particular \textit{Super Mario Bros.} levels. Plenty of work remains in applying these methods to other domains, including rulesets, items, characters, and 3-D levels. There is also very rapid progress within machine learning in general in the moment, and in particular within the deep learning field and in methods with generative capabilities such as Generative Adversarial Networks. There is plenty of interesting and rewarding work to do in exploring how these new capabilities can be adapted to function with the particular constraints and affordances of game content.

\ifCLASSOPTIONcaptionsoff
  \newpage
\fi



\bibliographystyle{IEEEtran}
\bibliography{bigliography}

%biographies
%%%%%%%%\begin{IEEEbiography}{Michael Shell}
%%%%%%%%Biography text here.
%%%%%%%%\end{IEEEbiography}
%%%%%%%%\begin{IEEEbiographynophoto}{John Doe}
%%%%%%%%Biography text here.
%%%%%%%%\end{IEEEbiographynophoto}
%\newpage
%%%%%%%%\begin{IEEEbiographynophoto}{Jane Doe}
%%%%%%%%Biography text here.
%%%%%%%%\end{IEEEbiographynophoto}
%\vfill
\end{document}